\documentclass{article}

\usepackage[preprint,nonatbib]{neurips_2023}
\usepackage[utf8]{inputenc} % allow utf-8 input
\usepackage[T1]{fontenc}    % use 8-bit T1 fonts
\usepackage[dvipsnames]{xcolor}         % colors
\definecolor{Klein_Blue}{rgb}{0.0, 0.129, 0.6}
\usepackage{hyperref}       % hyperlinks
\hypersetup{
    colorlinks=true,
    linkcolor=magenta,
    urlcolor=magenta,
    citecolor=Klein_Blue
    }
\usepackage{url}            % simple URL typesetting
\usepackage{booktabs}       % professional-quality tables
\usepackage{amsfonts}       % blackboard math symbols
\usepackage{nicefrac}       % compact symbols for 1/2,\etc
\usepackage{microtype}      % microtypography
\usepackage{xcolor, color, colortbl}         % colors
\usepackage{comment}
\usepackage{graphicx}
\usepackage{wrapfig}    % for warpping a figure with text 
\usepackage{tikz, siunitx}  % for darwing dash line
\usepackage{amsmath}    % math symbols
\usepackage{amssymb}    % for equations split
\usepackage{xspace}
\usepackage{tabularx}
\usepackage{threeparttable} % for table with footnote
\usepackage{rotating}
\usepackage{multirow}
\usepackage{multicol}
\usepackage{subcaption}
\usepackage{makecell}
\usepackage[export]{adjustbox}
\usepackage{enumitem} % itemize
\usepackage{gensymb} % degree
\usepackage{bm}

\newcommand{\figlabel}{Fig.\xspace}
\newcommand{\eqlabel}{Eq.\xspace}
\newcommand{\seclabel}{Sec.\xspace}

\newcommand{\tablabel}{Tab.\xspace}
\newcommand{\inlinesection}[1]{\noindent\textbf{#1.}}

\definecolor{Highlight}{HTML}{39b54a}  % green

\usepackage{amssymb}% http://ctan.org/pkg/amssymb
\usepackage{pifont}% http://ctan.org/pkg/pifont
\usepackage{algorithm}
\usepackage{listings}
\usepackage{etoolbox}
\makeatletter
\AfterEndEnvironment{algorithm}{\let\@algcomment\relax}
\AtEndEnvironment{algorithm}{\kern2pt\hrule\relax\vskip3pt\@algcomment}
\let\@algcomment\relax
\newcommand\algcomment[1]{\def\@algcomment{\footnotesize#1}}
\renewcommand\fs@ruled{\def\@fs@cfont{\bfseries}\let\@fs@capt\floatc@ruled
  \def\@fs@pre{\hrule height.8pt depth0pt \kern2pt}%
  \def\@fs@post{}%
  \def\@fs@mid{\kern2pt\hrule\kern2pt}%
  \let\@fs@iftopcapt\iftrue}
\makeatother
\newcommand{\cmmnt}[1]{}

\usepackage[normalem]{ulem}

\newcommand{\webpage}{\url{https://guochengqian.github.io/project/magic123}~}

\definecolor{mediumtealblue}{rgb}{0.0, 0.33, 0.71}
\definecolor{darkpastelgreen}{rgb}{0.01, 0.75, 0.24}
\definecolor{azure}{rgb}{0.0, 0.5, 1.0}
\newcommand{\colorA}[1]{\textcolor{Red}{#1}} % Note that the "Red" color is different from "red" :|
\newcommand{\colorC}[1]{\textcolor{azure}{#1}}
\newcommand{\colorB}[1]{\textcolor{darkpastelgreen}{#1}}
\newcommand{\methodname}{Magic\colorA{1}\colorB{2}\colorC{3}}

\begin{document}
%%%%%%%%% TITLE
\title{\methodname: \colorA{One} Image \colorB{to} High-Quality \colorC{3D} Object Generation Using Both 2D and 3D Diffusion Priors
}

% \title{Magic\textcolor{red}{1}\textcolor{blue}{2}\textcolor{green}{3}: \textcolor{red}{One} Image \textcolor{blue}{to} Realistic \textcolor{green}{3D} Object Generation}

\author{
Guocheng Qian$^{1,2}$, 
Jinjie Mai$^{1}$,
\textbf{Abdullah Hamdi$^{3}$},
Jian Ren$^{2}$, 
Aliaksandr Siarohin$^{2}$, 
\textbf{Bing Li}$^{1}$, \\
\textbf{Hsin-Ying Lee}$^{2}$, 
\textbf{Ivan Skorokhodov$^{1}$}, 
\textbf{Peter Wonka$^{1}$}, 
\textbf{Sergey Tulyakov$^{2}$}, 
\textbf{Bernard Ghanem$^{1}$}
\\
$^1$King Abdullah University of Science and Technology (KAUST), \quad $^2$Snap Inc.\\
$^3$Visual Geometry Group, University of Oxford\\
\texttt{\{guocheng.qian, bernard.ghanem\}@kaust.edu.sa}
% \\
% \texttt{\webpage}
}

\maketitle
\vspace{-2em}
\begin{figure}[H] 
\centering
\includegraphics[width=\textwidth]{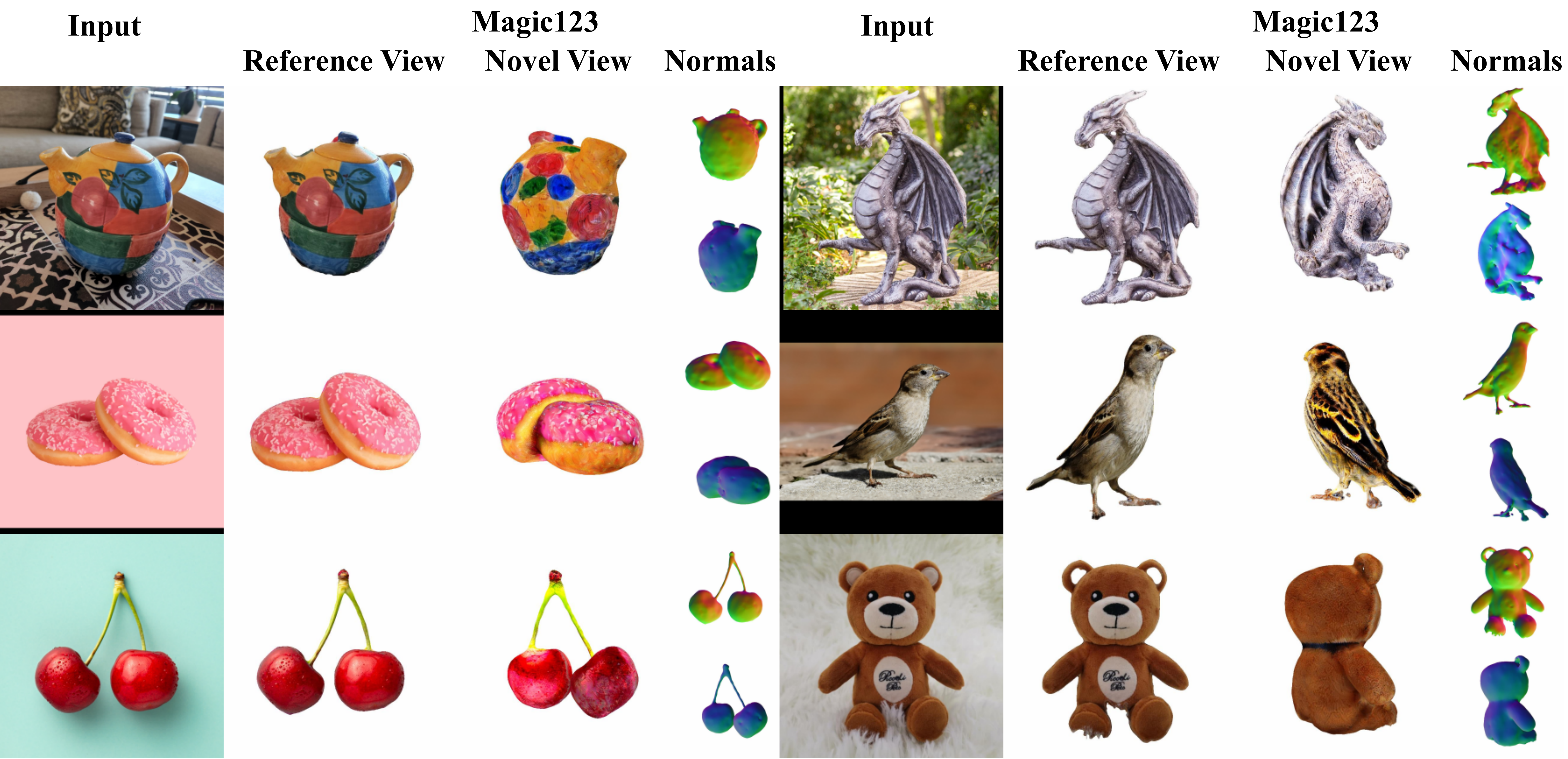}
\caption{\textbf{Magic123 for image-to-3D generation}. 
Magic123 can reconstruct high-fidelity 3D content with detailed 3D geometry and high rendering resolution ($1024\times1024$) from a \emph{single} unposed image in the wild. Visit \webpage for an immersive visualization. 
}
\label{fig:featured}
\end{figure}
\begin{abstract}
We present ``\textit{Magic123}'', a two-stage coarse-to-fine approach for high-quality, textured 3D meshes generation from a \textit{single unposed} image in the wild using \textit{both 2D and 3D priors}. In the first stage, we optimize a neural radiance field to produce a coarse geometry. In the second stage, we adopt a memory-efficient differentiable mesh representation to yield a high-resolution mesh with a visually appealing texture. In both stages, the 3D content is learned through reference view supervision and novel views guided by a combination of 2D and 3D diffusion priors. We introduce a single trade-off parameter between the 2D and 3D priors to control exploration (more imaginative) and exploitation (more precise) of the generated geometry. Additionally, we employ textual inversion and monocular depth regularization to encourage consistent appearances across views and to prevent degenerate solutions, respectively. Magic123 demonstrates a significant improvement over previous image-to-3D techniques, as validated through extensive experiments on synthetic benchmarks and diverse real-world images. Our code, models, and generated 3D assets are available at \url{https://github.com/guochengqian/Magic123}.
\end{abstract}
\section{Introduction}\label{sec:intro}

Despite observing the world in 2D, human beings have a remarkable capability to navigate, reason, and engage with their 3D surroundings. This points towards a deep-seated cognitive understanding of the characteristics and behaviors of the 3D world  - a truly impressive facet of human nature. This ability is taken to another level by artists who can produce detailed 3D replicas from a single image. Contrarily, from the perspective of computer vision, the task of 3D reconstruction from an unposed image - which encompasses the creation of geometry and textures - remains an unresolved, ill-posed problem, despite decades of exploration and development \cite{pixel2mesh,deepsdf,meshconv,Point2Mesh}.

The recent advances in deep learning \cite{GAN,StyleGAN,NeRF,LDM} have allowed an increasing number of 3D generation tasks to become learning-based. Even though deep learning has accomplished significant strides in image recognition \cite{resnet,vit} and generation \cite{GAN,StyleGAN,LDM}, the particular task of single-image 3D reconstruction in the wild is still lagging. We attribute this considerable discrepancy in 3D reconstruction abilities between humans and machines to two primary factors: (i) a deficiency in large-scale 3D datasets that impedes large-scale learning of 3D geometry, and (ii) the trade-off between the level of detail and computational resources when working on 3D data.

% 2d prior and its limitation
One possible approach to tackle the problem is to employ 2D priors. The pool of realistic 2D image data available online is voluminous. LAION \cite{LAION}, one of the most extensive text-image pair datasets, aids in training modern image understanding and generation models like CLIP~\cite{CLIP} and Stable Diffusion~\cite{LDM}. With the increasing generalization capabilities of 2D generation models, there has been a notable rise in approaches that use 2D models as priors for generating 3D content. DreamFusion~\cite{DreamFusion} serves as a trailblazer for this 2D prior-based methodology for text-to-3D generation. The technique demonstrates an exceptional capacity to guide novel views and optimize a neural radiance field (NeRF)~\cite{NeRF} in a zero-shot setting. Drawing upon DreamFusion, recent work such as RealFusion~\cite{RealFusion} and NeuralLift \cite{NeuralLift}, have endeavored to adapt these 2D priors for single image 3D reconstructions. 

\begin{figure}[hb] 
\centering
\includegraphics[width=\textwidth]{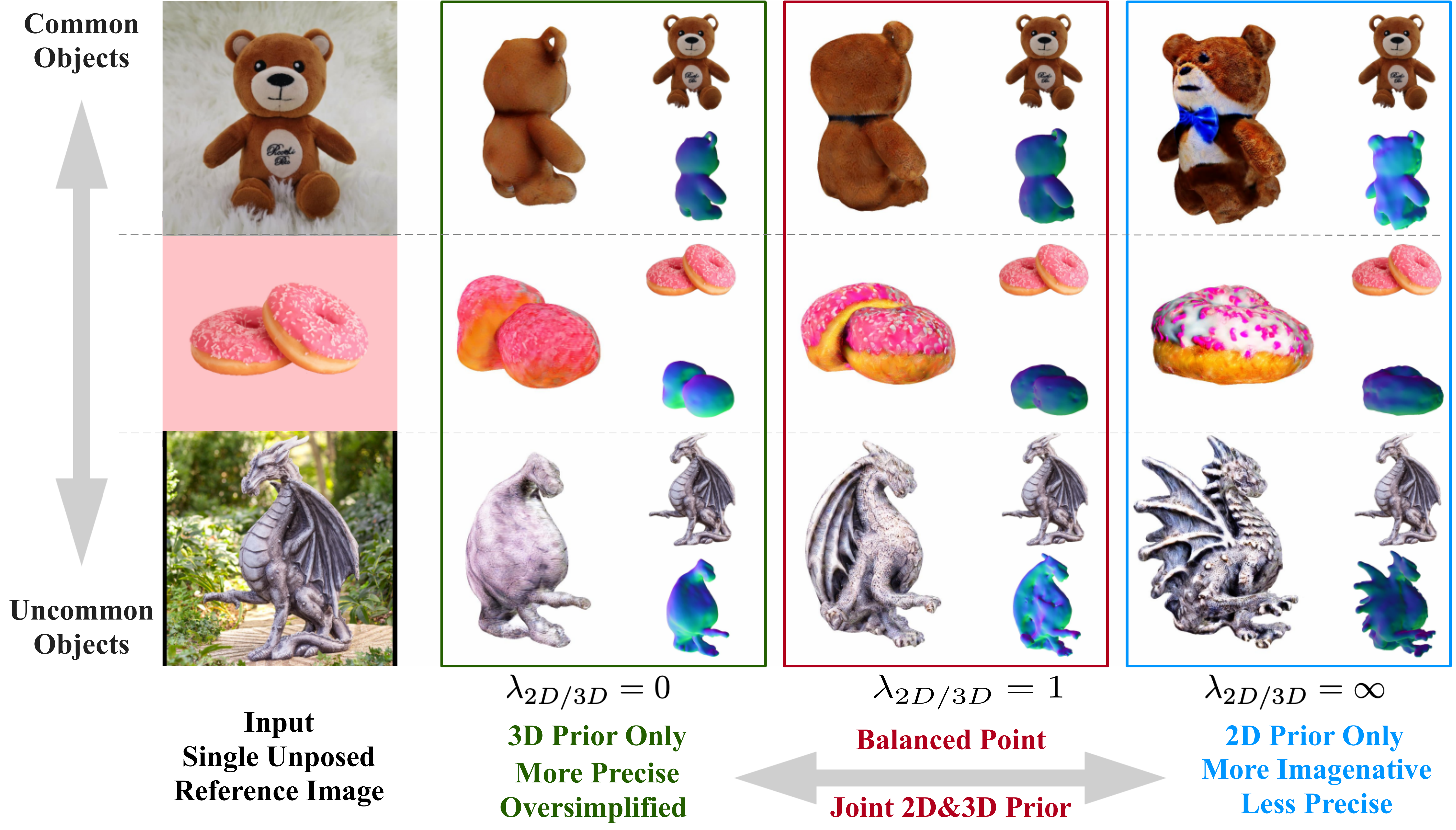}
\caption{
\textbf{Trade-off between 2D and 3D priors in Magic123}. We compare single image reconstructions for three cases: a teddy bear (common object), two stacked donuts (less common object), and a dragon statue (uncommon object).
As shown on the right, Magic123 with only a 2D prior favors geometry exploration, generating 3D content with more imagination while potentially lacking 3D consistency. Magic123 with only 3D prior (on the left) prioritizes geometry exploitation, resulting in precise yet potentially simplified geometry with reduced details. Magic123 thus proposes to use both 2D and 3D prior and introduces a trade-off parameter $\lambda_{2D/3D}$ to control the geometry exploration and exploitation (see \figlabel\ref{fig:lambda_2d-3d}). We provide a balanced point $\lambda_{2D/3D}$=1, with which Magic123 consistently offers identity-preserving 3D content with fine-grained geometry and visually appealing texture. 
}
\label{fig:pullfigure}
\end{figure}

% 3D Prior and its limitation 
Another approach is to employ 3D priors. Earlier attempts at 3D reconstruction leveraged 3D priors like topology constraints to assist in 3D generation \cite{pixel2mesh,text2mesh,deepsdf,sparf}. However, these manually-crafted 3D priors fall short of generating high-quality 3D content. Recently, approaches like Zero-1-to-3 \cite{Zero-1-to-3} and 3Dim \cite{3Dim} adapted a 2D diffusion model \cite{LDM} to become view-dependent and utilized this view-dependent diffusion as a 3D prior.

% Analyzing 2D and 3D priors
We analyzed the behavior of both 2D and 3D priors and found that they both have advantages and disadvantages. 2D priors exhibit impressive generalization for 3D generation that is unattainable with 3D priors (\emph{e.g.}, the dragon statue example in \figlabel{\ref{fig:pullfigure}). However, methods relying on 2D priors alone inevitably compromise on 3D fidelity and consistency due to their restricted 3D knowledge. This leads to unrealistic geometry like multiple faces (Janus problems), mismatched sizes, inconsistent texture, and so on. An instance of a failure case can be observed in the teddy bear example in \figlabel{\ref{fig:pullfigure}}.
On the other hand, a strict reliance on 3D priors alone is unsuitable for in-the-wild reconstruction due to the limited 3D training data. Consequently, as illustrated in \figlabel{\ref{fig:pullfigure}}, while 3D prior-based solution effectively processes common objects (for instance, the teddy bear example in the top row), it struggles with less common ones, yielding oversimplified, sometimes even flat 3D geometries (\emph{e.g.}, dragon statue at bottom left).

In this paper, rather than solely relying on a 2D or 3D prior, we advocate for the simultaneous use of both priors to guide novel views in image-to-3D generation. By modulating the simple yet effective tradeoff parameter between the potency of the 2D and 3D priors, we can manage the balance between exploration and exploitation in the generated 3D geometry. Prioritizing the 2D prior can enhance imaginative 3D capabilities to compensate for the incomplete 3D information inherent in a single 2D image, but this may result in less accurate 3D geometry due to a lack of 3D knowledge. In contrast, prioritizing the 3D prior can lead to more 3D-constrained solutions, generating more accurate 3D geometry, albeit with reduced imaginative capabilities and diminished ability to discover plausible solutions for challenging and uncommon cases.
We introduce Magic123, a novel image-to-3D pipeline that yields high-quality 3D outputs through a two-stage coarse-to-fine optimization process utilizing both 2D and 3D priors. In the coarse stage, we optimize a neural radiance field (NeRF)~\cite{NeRF}. NeRF learns an implicit volume representation, which is highly effective for complex geometry learning. However, NeRF demands significant memory, resulting in low-resolution rendered images passed to the diffusion models, making the output for the image-to-3D task low-quality.
Even the more resource-efficient NeRF alternative, Instant-NGP~\cite{InstantNGP}, can only reach a resolution of $128 \times 128$ in the image-to-3D pipeline on a 16GB memory GPU. Hence, to improve the quality of the 3D content, we introduce a second stage, employing a memory-efficient and texture-decomposed SDF-Mesh hybrid representation known as Deep Marching Tetrahedra (DMTet) \cite{DMTet}. This approach enables us to increase the resolution up to 1K and refine the geometry and texture of the NeRF separately.
In both stages, we leverage a combination of 2D and 3D priors to guide the novel views.

We summarize our contributions as follows:
\begin{itemize}[leftmargin=1em,topsep=0pt]
\item We introduce Magic123, a novel image-to-3D pipeline that uses a \textit{two-stage coarse-to-fine} optimization process to produce \emph{high-quality high-resolution} 3D geometry and textures.
\item We propose to use \textit{2D and 3D priors simultaneously} to generate faithful 3D content from any given image.  The strength parameter of priors allows for the trade-off between geometry exploration and exploitation. Users therefore can play with this trade-off parameter to generate desired 3D content.
\item Moreover, we find a balanced trade-off between 2D and 3D priors, leading to reasonably realistic and detailed 3D reconstructions. Using the exact \emph{same} set of parameters for all examples without any additional reconfiguration, Magic123 achieves state-of-the-art results in 3D reconstruction from single unposed images in both real-world and synthetic scenarios.
\end{itemize}
\section{Methodology}\label{sec:method}

\begin{figure}[t]
\centering
\includegraphics[page=1,width=1.0\textwidth, trim= 0 80cm 38cm 0, clip]{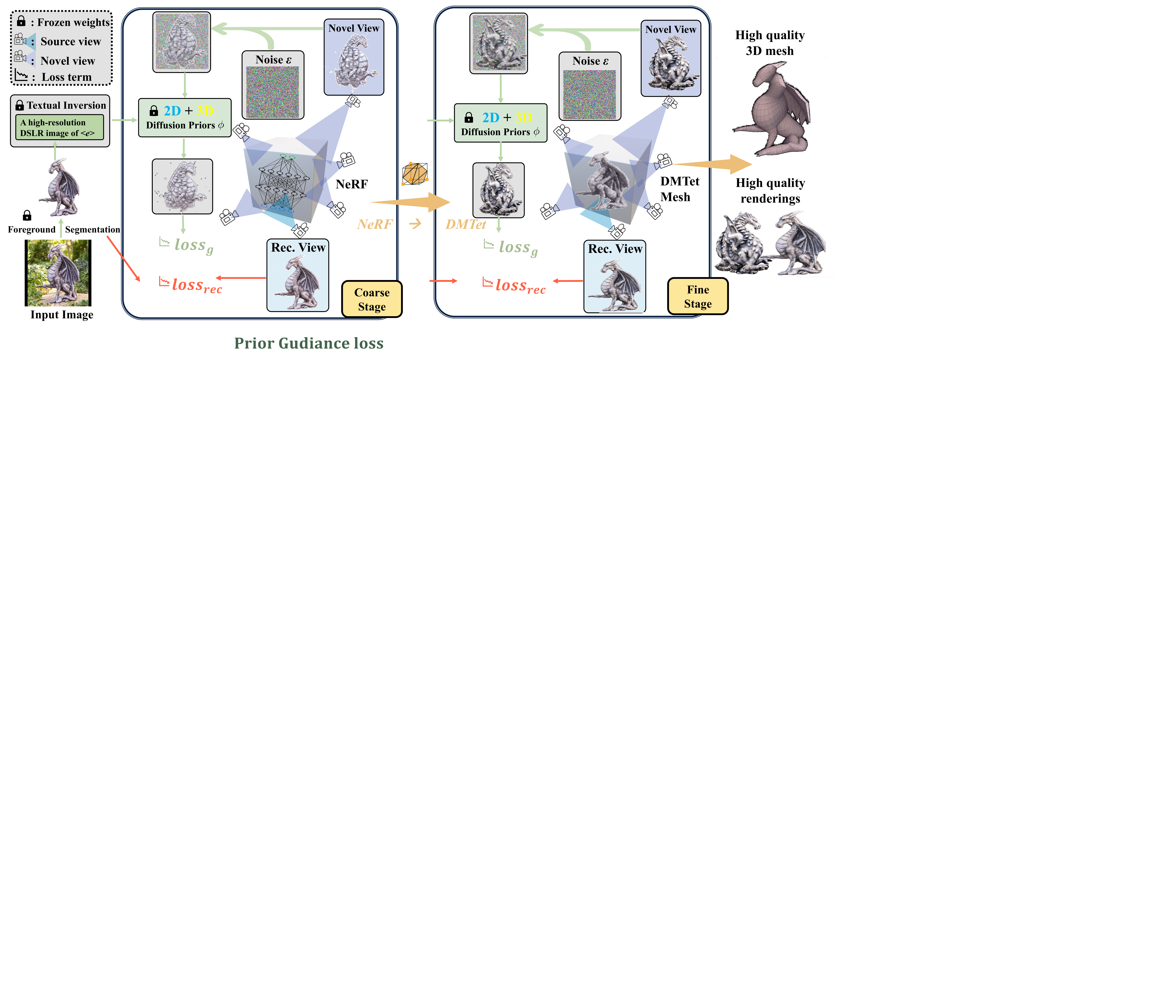}
\caption{\textbf{The pipeline of Magic123}. Magic123 is a two-stage coarse-to-fine framework for high-quality 3D generation from a reference image. Magic123 is guided by the reference image, constrained by the monocular depth estimation from the image, and driven by a joint 2D and 3D diffusion prior to dream up novel views. At the coarse stage, we optimize an Instant-NGP neural radiance field (NeRF) to reconstruct a coarse geometry. At the fine stage, we initialize a DMTet mesh from the NeRF output and optimize a high-resolution mesh and texture. Textural inversion is used in both stages to generate object-preserving geometry and view-consistent textures.
}
\label{fig:pipeline}
\end{figure}

We propose a two-stage framework, Magic123, that generates 3D content from a single reference image in a coarse to fine fashion, as shown in Fig.~\ref{fig:pipeline}. 
In the coarse stage, Magic123 learns a coarse geometry and texture by optimizing a NeRF. In the fine stage, Magic123 improves the quality of 3D content by directly optimizing a memory-efficient differentiable mesh representation with high-resolution renderings. In both stages, Magic123 uses joint 2D and 3D diffusion priors to trade off geometry exploration and geometry exploitation, yielding reliable 3D content with high generalizability. 

\subsection{Magic123 pipeline}

\inlinesection{Image preprocessing}
Magic123 is a pipeline for object-level image-to-3D generation. Given an image with a background, Magic123 requires a preprocessing step to extract the foreground object. We leverage an off-the-shelf segmentation model, Dense Prediction Transformer \cite{DPT}, to segment the object.
The extracted mask, denoted as
$\mathbf{M}$ is a binary segmentation mask and will be used in the optimization.
To prevent flat geometry collapse, \ie the model generates textures that only appear on the surface without capturing the actual geometric details, we further extract the depth map from the reference view by the pretrained MiDaS \cite{MiDaS}. 
The foreground image is used as the input, while the mask and the depth map are used in the optimization as regularization priors. 
These reference images are assigned fixed camera poses, assumed to the front view. More details in camera settings can be found in \seclabel{\ref{sec:details}}.

\subsubsection{Coarse stage} 

The coarse stage of our Magic123 is aimed at learning underlying geometry that respects the reference image. Due to its strong ability in handling complex topological changes in a smooth and continuous fashion, we adopt NeRF in this stage.

\inlinesection{Instant-NGP and its optimization} We leverage Instant-NGP \cite{InstantNGP} as our NeRF implementation because of its fast inference and ability to recover complex geometry. To reconstruct 3D faithfully from a single image, the optimization of NeRF requires at least two loss functions: (i) the reference view reconstruction supervision; and (ii) the novel view guidance.

\textbf{Reference view reconstruction loss} $\mathcal{L}_{rec}$ is imposed in our pipeline as one of the major loss functions to ensure the rendered image from the reference viewpoint ($\mathbf{v}^r$, assumed to be front view) is as close to the reference image $\mathbf{I}^r$ as possible. We adopt the mean squared error (MSE) loss on both the reference image and its mask as follows: 
\begin{equation}
\mathcal{L}_{rec}=\lambda_{rgb}\|\mathbf{M} \odot ( \mathbf{I}^r - G_\theta(\mathbf{v}^r))\|_2^2 + \lambda_{mask}\|\mathbf{M} - M(G_\theta(\mathbf{v}^r)))\|_2^2,
\end{equation}
where   
$\theta$ is the NeRF parameters to be optimized, $\odot$ is Hadamard product, 
$G_\theta(\mathbf{v}^r)$ is NeRF rendered view from $\mathbf{v}^r$ viewpoint, $M()$ is the foreground mask acquired by integrating the volume density along the ray of each pixel. Since the foreground object is extracted as input, we do not model any background and simply use pure white for the background rendering for all experiments. 
$\lambda_{rgb}, \lambda_{mask}$
are the weights for the foreground RGB and the mask.

\textbf{Novel view guidance $\mathcal{L}_{g}$} is necessary since multiple views are required to train a NeRF. 
We follow the pioneering work in text/image-to-3D \cite{DreamFusion,NeuralLift} and use diffusion priors to guide the novel view generation. As a significant difference from previous works, we do not rely solely on a 2D prior or a 3D prior, but we use both of them to guide the optimization of the NeRF. See \S\ref{sec:priors} for details.

\textbf{Depth prior} $\mathcal{L}_{d}$ is exploited to avoid overly-flat or caved-in 3D content. 
Using only the appearance reconstruction losses might yield poor geometry due to the inherent ambiguity of reconstructing 3D content from 2D images: the content of 3D may lie at any distance and still be rendered as the same image. This ambiguity might result in flat or curved-in geometry as noted in previous works \cite{NeuralLift}. We alleviate this issue by leveraging a depth regularization. 
% \WM{maybe add a citation? NeuralLift Sec 4.4 has discussed why depth is necessary}
A pretrained monocular depth estimator \cite{MiDaS} is leveraged to acquire the pseudo depth $d^r$ on the reference image. The depth output $d$ from the NeRF model from the reference viewpoint should be close to the depth prior. However, due to the value mismatch of two different sources of depth estimation, an MSE loss is not an ideal loss function. We use the normalized negative Pearson correlation as the depth regularization:
\begin{equation}
\mathcal{L}_{d}=\frac{1}{2}\left[1 - \frac{\text{cov}(\mathbf{M}\odot d^r, \mathbf{M}\odot d)}{\sigma(\mathbf{M}\odot d^r)\sigma(\mathbf{M}\odot d)}\right],
\end{equation}
where $\text{cov}(\cdot)$ denotes covariance  and $\sigma(\cdot)$ measures standard deviation.

\textbf{Normal smoothness} $\mathcal{L}_{n}$\textbf{.} 
One of the NeRF limitations is the tendency to produce high-frequency artifacts on the surface of the object. To this end, we enforce the smoothness of the normal maps of geometry for the generated 3D model following \cite{RealFusion}. We use the finite differences of the depth to estimate the normal vector of each point, render a 2D normal map $\mathbf{n}$ from the normal vector, and impose a loss as follows:
\begin{equation}
\mathcal{L}_{n}= \|\mathbf{n} - \tau(g(\mathbf{n},k)) \|,
\end{equation}
where $\tau(\cdot)$ denotes the stopgradient operation and $g(\cdot)$ is a Gaussian blur. The kernel size of the blurring $k$ is set to $9\times 9$.

Overall, the coarse stage is optimized by a combination of losses:
\begin{equation}\label{eqn:loss}
\mathcal{L}_{c}= \mathcal{L}_{rec} + \mathcal{L}_{g} + \lambda_d\mathcal{L}_d + \lambda_n\mathcal{L}_n ,
\end{equation}
where $\lambda_d, \lambda_n$ are the weights of depth and normal regularizations.

\subsubsection{Fine stage}\label{sec:fine}
The coarse stage offers a low-resolution 3D model, possibly with noise due to the tendency of NeRF to create high-frequency artifacts. 
Our fine stage aims to refine the 3D model and obtain a high-resolution and disentangled geometry and texture. To this end, we adopt DMTet \cite{DMTet}, which is a hybrid SDF-Mesh representation and is capable of generating high-resolution 3D shapes due to its high memory efficiency. 
Note the fine stage is identical to the coarse stage except for the 3D representation and rendering.

DMTet represents the 3D shape in terms of a deformable tetrahedral grid $(V_T, T)$, where $T$ denotes the tetrahedral grid and  $V_T$ are its vertexes. Given a vertex $v_i\in V_T$, a Signed Distance Function (SDF) $s_i \in \mathbb{R}$ and a triangle deformation vector $\triangle v_i\in \mathbb{R}^3$ are the parameters to be learned during optimization to extract a differentiable mesh \cite{DMTet}. 
The SDF is initialized by converting the density field of the coarse stage, while the triangle deformation is initialized as zero. 
For the textures, we follow Magic3D \cite{Magic3D} to use a neural color field that is initialized from the color field of the coarse stage. 
Since differentiable rasterization can be performed efficiently at very high resolution, we always use $8\times$ resolution of the coarse stage, which is found to have a similar memory consumption to the coarse stage.

\subsection{Joint 2D and 3D priors for image-to-3D generation}\label{sec:priors}

\inlinesection{2D priors}
Using a single reference image is insufficient to train a complete NeRF model without any priors \cite{PixelNeRF,visionnerf}.
To address this issue, DreamFusion \cite{DreamFusion} proposes to use a 2D diffusion model as the prior to guide the novel views via the proposed score distillation sampling (SDS) loss.
SDS exploits a 2D text-to-image diffusion model \cite{Imagen}, encodes the rendered view as latent, adds noise to it, and guesses the clean novel view guided by the input text prompt. Roughly speaking, SDS translates the rendered view into an image that respects both the content from the rendered view and the prompt. 
The SDS loss is illustrated in the upper part of \figlabel \ref{fig:2d_3d_prior} and is formulated as: 
\begin{equation}\label{eq:sds}
    \begin{aligned}
\mathcal{L}_{2D}=\mathbb{E}_{t, \mathbf{\epsilon}}\left[w(t)(\mathbf{\epsilon}_{\phi}(\mathbf{z}_{t};\mathbf{e},t)-\mathbf{\epsilon})\frac{\partial \mathbf{z}}{\partial \mathbf{I}}\frac{\partial \mathbf{I}}{\partial\theta}\right],
\end{aligned}
\end{equation}
where  $\mathbf{I}$ is a rendered view, and   $\mathbf{z}_t$ is the noisy latent by adding a random Gaussian noise of a time step $t$ to the latent of $\mathbf{I}$. $\epsilon, \epsilon_\phi$, $\phi$, $\theta$ are the added noise, predicted noise, parameters of the 2D diffusion prior, and the parameters of the 3D model. $\theta$ can be MLPs of NeRF for the coarse stage, or SDF, triangular deformations, and color field for the fine stage. 
DreamFusion \cite{DreamFusion} further points out that the Jacobian term of the image encoder $\frac{\partial \mathbf{z}}{\partial \mathbf{I}}$ in \eqlabel \eqref{eq:sds} can be further eliminated, making the SDS loss much more efficient in terms of both speed and memory.
In our experiments, we utilize the SDS loss with Stable Diffusion \cite{LDM} v1.5 as our 2D prior. The rendered images are interpolated to $512\times 512$ as required by the image encoder in \cite{LDM}.

\inlinesection{Textural inversion}
Note the prompt $\mathbf{e}$ we use for each reference image is not a pure text chosen from tedious prompt engineering. Using pure text for image-to-3D generation most likely results in inconsistent texture and geometry due to the limited expressiveness of the human language. For example, using ``A high-resolution DSLR image of a colorful teapot'' will generate different geometry and colors that do not respect the reference image. We thus follow RealFusion \cite{RealFusion} to leverage the same textual inversion \cite{TextualInversion} technique to acquire a special token \textit{<e>} to represent the object in the reference image. 
We use the same prompt for all examples: ``A high-resolution DSLR image of \textit{<e>}''. 
We find that Stable Diffusion can generate the teapot with a more similar texture and style to the reference image with the textural inversion technique compared to the results without it.
% (see \supp). 

Overall, the 2D diffusion priors \cite{DreamFusion,Latent-NeRF,Magic3D} exhibit a remarkable capacity for exploring the space of geometry, thereby facilitating the generation of diverse geometric representations with a heightened sense of imagination. This exceptional imaginative capability compensates for the inherent limitations associated with the availability of incomplete 3D information in a single 2D image. Moreover, the utilization of 2D prior-based techniques for 3D reconstruction reduces the likelihood of overfitting in certain scenarios, owing to their training on an extensive dataset comprising over a billion images. However, it is crucial to acknowledge that the reliance on 2D priors may introduce inaccuracies in the generated 3D representations, thereby potentially deviating from true fidelity. This low-fidelity generation happens because 2D priors lack 3D knowledge. For instance, the utilization of 2D priors may yield imprecise geometries, such as Janus problems and mismatched sizes as depicted in \figlabel~\ref{fig:pullfigure} and \figlabel~\ref{fig:lambda_2d-3d}.

\begin{figure}[t] 
\centering
\includegraphics[page=2,width=0.78\textwidth, trim= 0 106.5cm 93cm 0, clip]{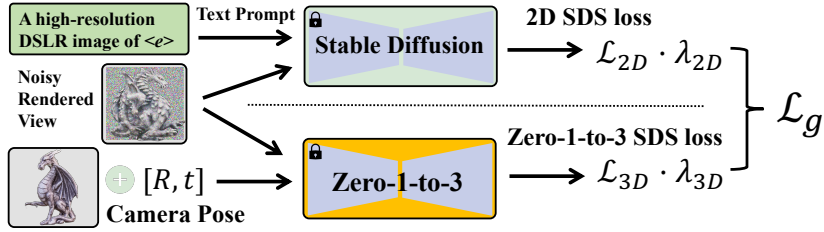}
\caption{\textbf{2D \vs 3D Diffusion priors.} 
Magic123 uses Stable Diffusion \cite{LDM} as the 2D prior and viewpoint-conditioned diffusion model Zero-1-to-3~\cite{Zero-1-to-3} as the 3D prior. Stable Diffusion takes the noisy rendered view and a text prompt as input, while 
Zero-1-to-3 uses additionally the novel view camera pose as input, creating a 3D-aware prior for Magic123.
}
\label{fig:2d_3d_prior}
\end{figure}

\subsubsection{3D prior}
Using only the 2D prior is not sufficient to capture detailed and consistent 3D geometry. Zero-1-to-3 \cite{Zero-1-to-3} thus proposes a 3D prior solution. 
Zero-1-to-3 finetunes Stable Diffusion into a view-dependent version on Objaverse \cite{Objaverse}, the largest open-source 3D dataset that consists of 818K models. Zero-1-to-3 takes a reference image and a viewpoint as input and can generate a novel view from the given viewpoint. Zero-1-to-3 thereby can be used as a strong 3D prior for 3D reconstruction. 
The usage of Zero-1-to-3 in an image-to-3D generation pipeline using SDS loss \cite{DreamFusion} is formulated as: 
\begin{equation}\label{eq:3d}
    \begin{aligned}
\mathcal{L}_{3D}=\mathbb{E}_{t, \mathbf{\epsilon}}\left[w(t)(\mathbf{\epsilon}_{\phi}(\mathbf{z}_{t};\mathbf{I}^r, t, R, T)-\mathbf{\epsilon})\frac{\partial \mathbf{I}}{\partial\theta}\right],
\end{aligned}
\end{equation}
where $R, T$ are the camera poses passed to Zero-1-to-3, the view-dependent diffusion model.
The difference between using the 3D prior and the 2D prior is illustrated in \figlabel \ref{fig:2d_3d_prior}, where we show that the 2D prior uses text embedding as guidance while the 3D prior uses the reference view $\mathbf{I}^r$ with the novel view camera poses as guidance. The 3D prior utilizes camera poses to encourage 3D consistency and enable the usage of more 3D information compared to the 2D prior. 
% Note that other 3D priors available are Point-E \cite{PointE} and Shap-E \cite{ShapeE} which utilize a 3D dataset to train 3D generators from a single image. We leave the usage of Point-E and Shap-E as future work \todo{Shap-E in appendix}.

Overall, the utilization of 3D priors demonstrates a commendable capacity for effectively harnessing the expansive realm of geometry, resulting in the generation of significantly more accurate geometric representations compared to their 2D counterparts. This heightened precision particularly applies when dealing with objects that are commonly encountered within the pre-trained 3D dataset. However, it is essential to acknowledge that the generalization capability of 3D priors is comparatively lower than that of 2D priors, thereby potentially leading to the production of geometric structures that may appear implausible. This low generalization results from the limited scale of available 3D datasets, especially in the case of high-quality real-scanned objects. 
For instance, in the case of uncommon objects, the employment of Zero-1-to-3 often tends to yield overly simplified geometries, \eg flat surfaces without details in the back view (see \figlabel\ref{fig:pullfigure} and \figlabel\ref{fig:lambda_2d-3d}).

\subsubsection{Joint 2D and 3D priors}
We find that the 2D and 3D priors are complementary to each other. Instead of relying solely on 2D or 3D prior, we propose to use both priors in 3D generation.
The 2D prior is used to \textit{explore} the geometry space, favoring high imagination but might lead to inaccurate geometry. We name this characteristic of the 2D prior as \textit{geometry exploration}. 
On the other hand, the 3D prior is used to \textit{exploit} the geometry space, constraining the generated 3D content to fulfill the implicit requirement of the underlying geometry, favoring precise geometry but with less generalizability. In the case of uncommon objects, the 3D prior might result in over-simplified geometry. We name this feature of using the 3D prior as \textit{geometry exploitation}. In our image-to-3D pipeline, we propose a new prior loss for the novel view supervision to combine both 2D and 3D priors:
\begin{equation}\label{eq:joint}
    \begin{aligned}
\mathcal{L}_{g}=\mathbb{E}_{t_1,t_2, \mathbf{\epsilon}_1,\mathbf{\epsilon}_2}\left[w(t)\left[\lambda_{2D/3D}(\mathbf{\epsilon}_{\phi_{2D}}(\mathbf{z}_{t_1};\mathbf{e},t_1)-\mathbf{\epsilon}_1)+\lambda_{3D}(\mathbf{\epsilon}_{\phi_{3D}}(\mathbf{z}_{t_2};\mathbf{I}^r, t_2, R, T)-\mathbf{\epsilon}_2)\right]\frac{\partial \mathbf{I}}{\partial\theta}\right],
\end{aligned}
\end{equation}
where $\lambda_{2D/3D}$ and $\lambda_{3D}$ determine the strength of 2D and 3D prior, respectively. 
Weighting more on $\lambda_{2D/3D}$ leads to more geometry exploration, while weighting more on $\lambda_{3D}$ results in more geometry exploitation. However, tuning two parameters at the same time is not user-friendly. Interestingly, 
through both qualitative and quantitative experiments, we find that Zero-1-to-3, the 3D prior we use, is much more tolerant to $\lambda_{3D}$ than Stable Diffusion to $\lambda_{2D}$. When only the 3D prior is used, \ie $\lambda_{2D}=0$,  Zero-1-to-3 generates consistent results for $\lambda_{3D}$ ranging from 10 to 60. On the contrary, Stable Diffusion is rather sensitive to $\lambda_{2D}$. When setting $\lambda_{3D}$ to $0$ and using the 2D prior only, the generated geometry varies a lot when $\lambda_{2D}$ is changed from $1$ to $2$. 
This observation leads us to fix $\lambda_{3D}=40$ and to rely on tuning the $\lambda_{2D}$ to trade off the geometry exploration and exploitation.  We set $\lambda_{2D/3D}=1.0$ for all results throughout the paper, but this value can be tuned according to the user's preference. More details and discussions on the choice of 2D and 3D priors weights are available in \seclabel{\ref{sec:ablation}}.

\section{Experiments}\label{sec:exp}
\subsection{Datasets}\label{sec:dataset}

\inlinesection{NeRF4} 
We introduce a NeRF4 dataset that we collect from 4 scenarios, chair, drums, ficus, and microphone, out of the 8 test examples from the synthetic NeRF dataset \cite{NeRF}. 
These four scenarios cover complex objects (drums and ficus), a hard case (the back view of the chair), and a simple case (the microphone). The other four examples are removed since they are not subject to the front view assumption, requiring further camera pose estimation or a manual tuning of the camera pose, which is out of the scope of this work.
% \Refer to failure cases in \supp for details.

\inlinesection{RealFusion15} We further use the dataset collected and released by RealFusion \cite{RealFusion}, consisting of 15 natural images that include bananas, birds, cacti, barbie cakes, cat statues, teapots, microphones, dragon statues, fishes, cherries, and watercolor paintings \etc.

\begin{figure}[H] 
\centering
\includegraphics[width=\textwidth]{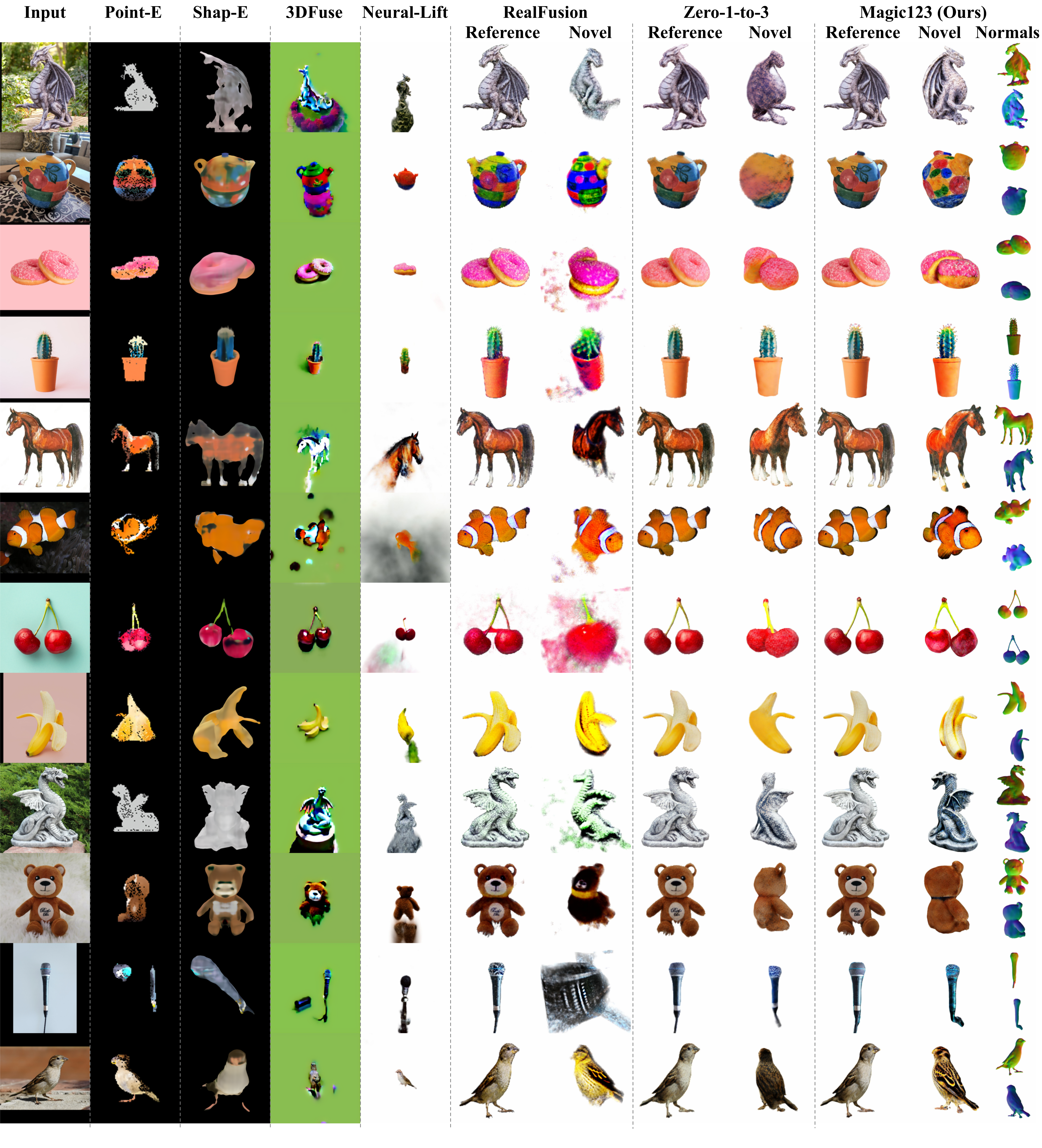}
% \vspace{1em}
\includegraphics[width=\textwidth]{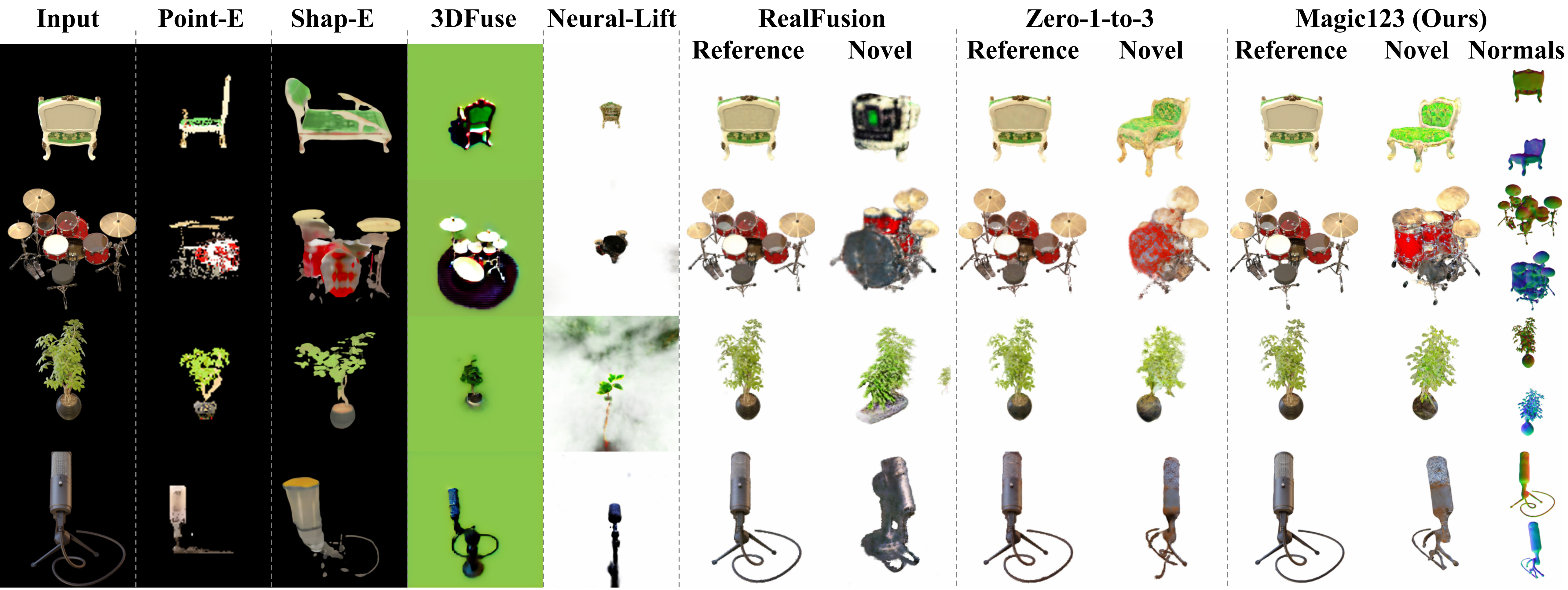}
\caption{
{\textbf{Qualitative comparisons on image-to-3D generation.} We compare Magic123 to recent methods (Point-E \cite{PointE}, ShapeE \cite{ShapeE}, 3DFuse \cite{3DFuse}, RealFusion \cite{RealFusion}, and Zero-1-to-3 \cite{Zero-1-to-3}) for generating 3D objects from a single unposed image (the leftmost column)}. On top, we show results on the RealFusion15 dataset, and on the bottom, we show results on the NeRF4 dataset.
}
\label{fig:sota1}
\end{figure}

\subsection{Implementation details} \label{sec:details}
\inlinesection{Optimizing the pipeline} 
We use \emph{exactly the same} set of hyperparameters for all experiments and do not perform any per-object hyperparameter optimization. 
Both coarse and fine stages are optimized using Adam with $0.001$ learning rate and no weight decay for $5,000$ iterations. 
$\lambda_{rgb}, \lambda_{mask}, \lambda_{d}$ are set to $5, 0.5, 0.001$ for both stages. $\lambda_{2D}$ and $\lambda_{3D}$ are set to $1$ and $40$ for the first stage and are lowered to $0.001$ and $0.01$ in the second stage for refinement to alleviate oversaturated textures.
We adopt the Stable Diffusion~\cite{DiffusionModels} model of V1.5 as the 2D prior. The guidance scale of the 2D prior is set to $100$ following \cite{DreamFusion}. 
For the 3D prior, Zero-1-to-3 \cite{Zero-1-to-3} ($105,000$ iterations finetuned version) is leveraged. The guidance scale of Zero-1-to-3 is set to $5$ following \cite{Zero-1-to-3}.
The NeRF backbone is implemented by three layers of multi-layer perceptrons with $64$ hidden dims.
Regarding lighting and shading, we keep nearly the same as \cite{DreamFusion}. The difference is we set the first $3,000$ iterations in the first stage to normals' shading to focus on learning geometry. For other iterations as well as the fine stage, we use diffuse shading with a probability $0.75$ and textureless shading with a probability $0.25$.
The rendering resolutions are set to $128\times 128$ and $1024\times 1024$
for the coarse and the fine stage, respectively.

\inlinesection{Camera setting} 
Since the reference image is unposed, we assume its camera parameters are as follows.
First, the reference image is assumed to be shot from the front view, \ie polar angle $90\degree$, azimuth angle $0\degree$. 
Second, the camera is placed $1.8$ meters from the coordinate origin, \ie the radial distance is $1.8$.
Third, the field of view (FOV) of the camera is 40$\degree$. 
We highlight that the 3D reconstruction performance is not sensitive to camera parameters, as long as they are reasonable, \eg FOV between $20$ and $60$, and radial distance between $1$ to $4$ meters. 
Note this camera setting works for images subject to the front-view assumption. For images taken deviating from the front view, a manual change of polar angle or a camera estimation is required.
% See more details in \supp. 

\subsection{Results}\label{sec:results}
\inlinesection{Evaluation metrics}
For a comprehensive evaluation, we adhere to the metrics employed in prior studies~\cite{NeuralLift, RealFusion}, namely PSNR, LPIPS~\cite{LPIPS}, and CLIP-similarity~\cite{CLIP}. PSNR and LPIPS are gauged in the reference view to measure reconstruction quality and perceptual similarity. CLIP-similarity calculates an average CLIP distance between rendered image and the reference image to measure 3D consistency through appearance similarity across novel views and the reference view.

\inlinesection{Quantitative and qualitative comparisons}
We compare Magic123 against the state-of-the-art PointE \cite{PointE}, Shap-E \cite{ShapeE}, 3DFuse \cite{3DFuse}, NeuralLift \cite{NeuralLift}, RealFusion \cite{RealFusion} and Zero-1-to-3 \cite{Zero-1-to-3} in both NeRF4 and RealFusion15 datasets. 
For Zero-1-to-3, we adopt the implementation here \cite{stable-dreamfusion}, which yields better performance than the original implementation. For other works, we use their officially released code.   
As shown in Table~\ref{table:Image-to-3D}, Magic123 achieves Top-1 performance across all the metrics in both datasets when compared to previous approaches. It is worth noting that the PSNR and LPIPS results demonstrate significant improvements over the baselines, highlighting the exceptional reconstruction performance of Magic123. The improvement of CLIP-Similarity reflects the great 3D coherency regards to the reference view.
Qualitative comparisons are available in Fig. \ref{fig:sota1}.
Magic123 achieves the best results in terms of both geometry and texture. Note how Magic123 greatly outperforms the 3D-based zero-1-to-3 \cite{Zero-1-to-3} especially in complex objects like the dragon statue and the colorful teapot in the first two rows, while at the same time greatly outperforming 2D-based RealFusion \cite{RealFusion} in all examples. 
This performance demonstrates the superiority of Magic123 over the state-of-the-art and its ability to generate high-quality 3D content.

\begin{table}[h]
\centering
\caption{\textbf{Magic123 results.} We show quantitative results in terms of CLIP-Similarity$\uparrow$ / PSNR$\uparrow$ / LPIPS$\downarrow$. The results are shown on the NeRF4 and Realfusion datasets, while \textbf{bold} reflects the best.}
\label{table:Image-to-3D}
\resizebox{\textwidth}{!}{%
\begin{tabular}{@{}c|cccccccc@{}}
\toprule
\textbf{Dataset} & \textbf{Metrics\textbackslash{}Methods} 
& Point-E~\cite{PointE} & Shap-E~\cite{ShapeE} 
& 3DFuse~\cite{3DFuse} & NeuralLift~\cite{NeuralLift}  & RealFusion~\cite{RealFusion}
& Zero-1-to-3~\cite{Zero-1-to-3} & \textbf{Magic123} (\textbf{Ours})\\
\midrule
\multirow{3}{*}{\textbf{NeRF4}}        
& CLIP-Similarity$\uparrow$  & 0.48 & 0.60 & 0.60 & 0.52 & 0.38 & 0.62 & \textbf{0.80} \\
& PSNR$\uparrow$             & 0.70 & 0.99 & 5.86 & 12.55 & 15.37 & 23.96 & \textbf{24.62} \\
& LPIPS$\downarrow$           & 0.80 & 0.76 & 0.76 & 0.50 & 0.20 & 0.05 & \textbf{0.03} \\ \midrule
\multirow{3}{*}{\textbf{RealFusion15}} 
& CLIP-Similarity$\uparrow$ & 0.53 & 0.59 & 6.28 & 0.65 & 0.67 & 0.75 & \textbf{0.82} \\
& PSNR$\uparrow$            & 0.98 & 1.23 & 18.87 & 11.08 & 0.67 & 19.49 & \textbf{19.50} \\
& LPIPS$\downarrow$          & 0.78 & 0.74 & 0.80 & 0.53 & 0.14 & 0.11 & \textbf{0.10} \\ 
\bottomrule
\end{tabular}
}
\end{table}

% \begin{table}[t]
% \centering
% \caption{Quantitative results in terms of CLIP-Similarity$\uparrow$ / PSNR$\uparrow$ / LPIPS$\downarrow$. \textbf{Bold} reflects the best.}
% \label{table:Image-to-3D}
% \resizebox{\textwidth}{!}{%
% \begin{tabular}{lccccccc}
% \toprule
% \textbf{Dataset} & Point-E~\cite{PointE} & Shap-E~\cite{ShapeE} 
% & 3DFuse~\cite{3DFuse} & NeuralLift~\cite{NeuralLift}  & RealFusion~\cite{RealFusion}
% & Zero-1-to-3~\cite{Zero-1-to-3} & \textbf{Magic123} (\textbf{Ours}) \\
% \midrule
% NeRF4 & 0.48 / 0.70 / 0.80 & 0.60 / 0.99 / 0.76 & 0.60 / 5.86 / 0.76 & 0.52 / 12.55 / 0.50 & 0.38 / 15.37 / 0.20& 0.62 / 23.96 / 0.05 & \textbf{0.80 / 24.76 / 0.03}
% \\
% RealFusion15 & 0.53 / 0.98 / 0.78 & 0.59 / 1.23 / 0.74 & 
% 0.67 / 6.28 / 0.80 & 0.65 / 11.08 / 0.53 & 0.67 / 18.87 / 0.14 & 0.75 / 19.49 / 0.11 & \textbf{0.81 / 19.63 / 0.09} \\
% \bottomrule
% \end{tabular}
% }
% \end{table}

\subsection{Ablation and analysis}\label{sec:ablation}
Magic123 introduces a coarse-to-fine pipeline for single image reconstruction and a joint 2D and 3D prior for novel view guidance. 
We provide analysis and ablation studies to show their effectiveness.
% Due to the limited space, we leave the discussions of this part and quantitative results in \supp. 
% To further show the effectiveness of our method, we provide more quantitative analysis and conduct ablation studies. 
% We mainly show: (1) the coarse-to-fine strategy is essential for the high-quality output; (2) the output from the coarse stage (which shows great geometry but in low resolution); 
% (3) the geometry transition by tuning $\lambda_{div}$, where we show the geometry is going from a precise but over-smoothed shape ($\lambda_{div}=0$) to a diverse but less faithful shape ($\lambda_{div}=2$).

% \input{tables/ablation_2d-3d-coarse-fine}
\begin{figure}[b] 
\centering
\includegraphics[width=1.0\textwidth]{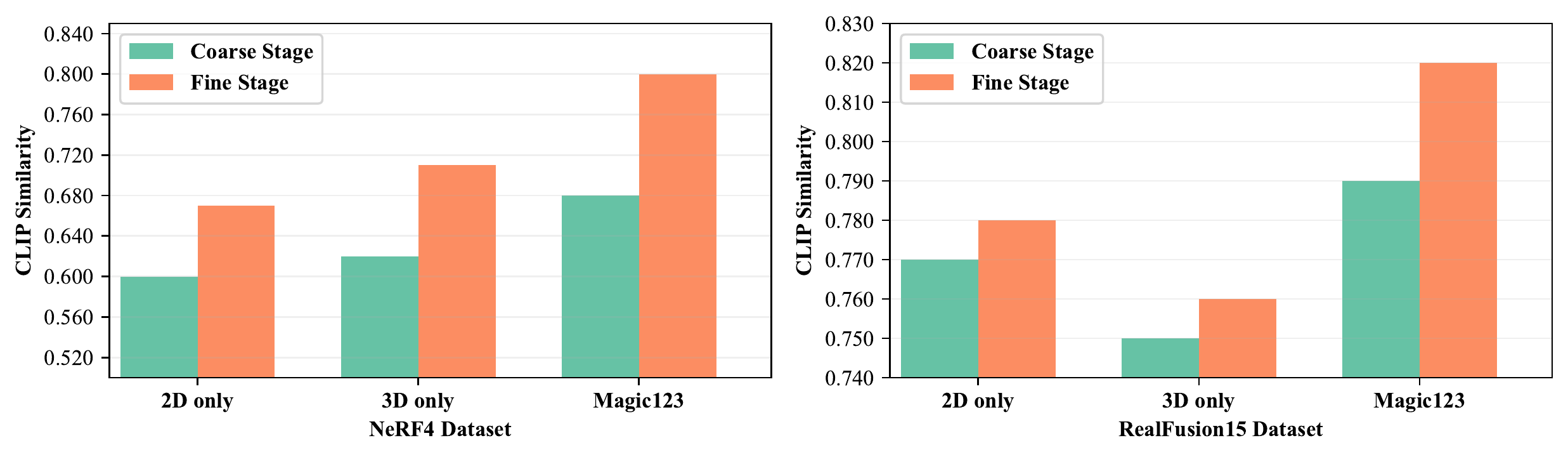} \\
% \includegraphics[width=\textwidth]{figures/src/2d-3d-coarse-fine.pdf}
% \vspace{-2em}
\caption{\textbf{Ablation study (quantitative).} We quantitatively compare using the coarse and fine stages in Magic123. In both setups, we ablate utilizing only 2D prior ($\lambda_{2D}$=$1$,$\lambda_{3D}$=$0$), utilizing only 3D prior ($\lambda_{2D}$=$0$,$\lambda_{3D}$=$40$), and utilizing both 2D and 3D priors ($\lambda_{2D}$=$1$,$\lambda_{3D}$=$40$).
% \todo{@wayne: lighter gridlines}
}
\label{fig:coarse-fine-plot}
\end{figure}

% \begin{table}[t]
% \centering
% \caption{Quantitative comparisons of the coarse and fine stages between Magic123 with only 2D prior ($\lambda_{2D}$=$1$,$=\lambda_{3D}$=$0$), Magic123 with only 3D prior $\lambda_{2D}$=$0$,$\lambda_{3D}$=$40$, and Magic123 ($\lambda_{2D}$=$1$,$\lambda_{3D}$=$40$)
% % \todo{@wayne update table, value comment below}
% }
% \label{tab:2d-3d-coarse-fine}
% \resizebox{\textwidth}{!}{%
% \begin{tabular}{@{}c|cccccccc@{}}
% \toprule
% \textbf{Dataset} & \textbf{Metrics\textbackslash{}Methods} 
% & 2D prior (Coarse) & 2D prior & 3D prior (Coarse) & 3D  prior & Magic123 Coarse & \textbf{Magic123}  \\ \midrule
% \multirow{3}{*}{\textbf{NeRF4}}        
% & CLIP-Similarity$\uparrow$  & 0.60 & 0.67 & 0.62 & 0.71 & 0.68  & \textbf{0.80} \\
% & PSNR$\uparrow$             & 24.11 & 24.58 & 23.96 & 24.69 & 23.67  & \textbf{24.62} \\
% & LPIPS$\downarrow$           & 0.04 & 0.03 & 0.05 & 0.03 & 0.05  & \textbf{0.03} \\ \midrule
% \multirow{3}{*}{\textbf{RealFusion15}} 
% & CLIP-Similarity$\uparrow$ & 0.77 & 0.78 & 0.75 & 0.76 & 0.79  & \textbf{0.82} \\
% & PSNR$\uparrow$            & 19.77 & 19.60 & 19.49 & 19.58 & 19.42  & \textbf{19.50} \\
% & LPIPS$\downarrow$          & 0.10 & 0.09 & 0.11 & 0.10 & 0.11  & \textbf{0.10} \\ \bottomrule
% \end{tabular}
% }
% \end{table}

\inlinesection{The effect of two stages}
We study in Fig. \ref{fig:coarse-fine-plot} and   Fig. \ref{fig:2d-3d-coarse-fine} the effect of using the fine stage of our pipeline on the performance of Magic123. We note that a consistent improvement in terms of both qualitative and quantitative performance is observed throughout different setups when the fine stage is combined with the coarse stage. The use of a textured mesh DMTet representation enables higher quality 3D content that fits the objective and produces more compelling and higher resolution 3D consistent visuals.

\inlinesection{3D priors only}
We first turn off the guidance of the 2D prior by setting $\lambda_{2D}=0$, such that we only use the 3D prior Zero-1-to-3 \cite{Zero-1-to-3} as the guidance. We study the effects of $\lambda_{3D}$ by setting it to $10, 20, 40, 60$. Interestingly, we find that Zero-1-to-3 is very robust to the change of $\lambda_{3D}$. \tablabel \ref{tab:ablate_2d_3d} demonstrates that different $\lambda_{3D}$ lead to a consistent quantitative result. We thus simply set $\lambda_{3D}=40$ throughout the experiments since it achieves a slightly better CLIP-similarity score than other values.

% \usepackage{graphicx}
\begin{table}[t]
\centering
\caption{\textbf{Effects} of  $\lambda_{3D}$ and  $\lambda_{2D}$ in Magic123 using only 2D or 3D prior on NeRF4 dataset.}
\label{tab:ablate_2d_3d}
\resizebox{0.85\textwidth}{!}{%
\begin{tabular}{c|ccccc|ccc}
\toprule
 & \multicolumn{5}{c|}{varying $\lambda_{3D}$ when $\lambda_{2D}$=$0$} & \multicolumn{3}{c}{varying $\lambda_{2D}$ when $\lambda_{3D}$=$0$} \\
 & \textit{10}  & \textit{20} & \textit{40} & \textit{60} & \textit{80} & \textit{0.1}       & \textit{1}       & \textit{2}      \\ \midrule
CLIP-similarity$\uparrow$ & 0.58  & 0.61  & 0.62  & 0.61  & 0.58  & 0.54  & 0.60  & 0.72  \\
PSNR$\uparrow$           & 23.96 & 24.05 & 23.96 & 23.75 & 23.34 & 23.62 & 24.11 & 22.42 \\
LPIPS$\downarrow$         & 0.04  & 0.04  & 0.05  & 0.06  & 0.08  & 0.04  & 0.04  & 0.07  \\ \bottomrule
\end{tabular}%
}
\end{table}

\inlinesection{2D priors only}
We then turn off the 3D prior and study the effect of $\lambda_{2D}$ in the image-to-3D task. As shown in \tablabel \ref{tab:ablate_2d_3d}, with the increase of $\lambda_{2D}$, an increase in CLIP-similarity is observed. This is due to the fact that a larger 2D prior weight leads to more imagination but unfortunately might result in the Janus problem.

\inlinesection{Combining both 2D and 3D priors and the trade off factor $\lambda_{2D/3D}$}
In Magic123, we propose to use both 2D and 3D priors.
% Our novel-view guidance loss is formulated as: $\lambda_{2D/3D} \mathcal{L}_{2D} + 40\mathcal{L}_{3D}$, where the trade-off parameter $\lambda_{2D/3D}$ is used to balance between the geometry exploitation by the 3D prior and the geometry exploration by the 2D prior.
\figlabel\ref{fig:coarse-fine-plot} demonstrates the effectiveness of combining the 2D and 3D priors on the quantitative performance of image-to-3D generation. 
% In \figlabel\ref{fig:2d-3d-coarse-fine} we show that (i) using 2D only prior achieves much better performance in imagining complex scenes such as the dragon statue, but also fails badly in the simple microphone examples; (ii) using 3D only prior achieves more precise geometry but fails in complex and uncommon objects (the drums), generating over-simplified back view geometry; (iii) using both 2D and 3D priors achieves a good tradeoff that can handle both simple and complex objects.
In \figlabel\ref{fig:lambda_2d-3d}, we further analyze the tradeoff hyperparameter $\lambda_{2D/3D}$ from \eqlabel\eqref{eq:joint}. We start from $\lambda_{2D/3D}$=$0$ to use only the 3D prior and gradually increase $\lambda_{2D/3D}$ to $0.1,0.5,1.0,2,5$, and finally $\infty$ to use only the 2D prior with $\lambda_{2D}$=$1$ and $\lambda_{3D}$=$0$. The key observations include:
(1) Relying solely on the 3D prior results in precise geometry (as observed in the teddy bear) but falters in generating complex and uncommon objects, often rendering oversimplified geometry with minimal details (as seen in the dragon statue);
(2) Relying solely on the 2D prior significantly improves performance in conjuring complex scenes like the dragon statue but simultaneously triggers the Janus problem in simple examples such as the bear;
(3) As $\lambda_{2D/3D}$ escalates, the imaginative prowess of Magic123 is enhanced and more details become evident, but there is a tendency to compromise 3D consistency. We assign $\lambda_{2D/3D}$=$1$ as the default value for all examples. However, this parameter could also be fine-tuned for even better results on certain inputs.

\begin{figure}[t] 
\centering
\includegraphics[width=\textwidth]{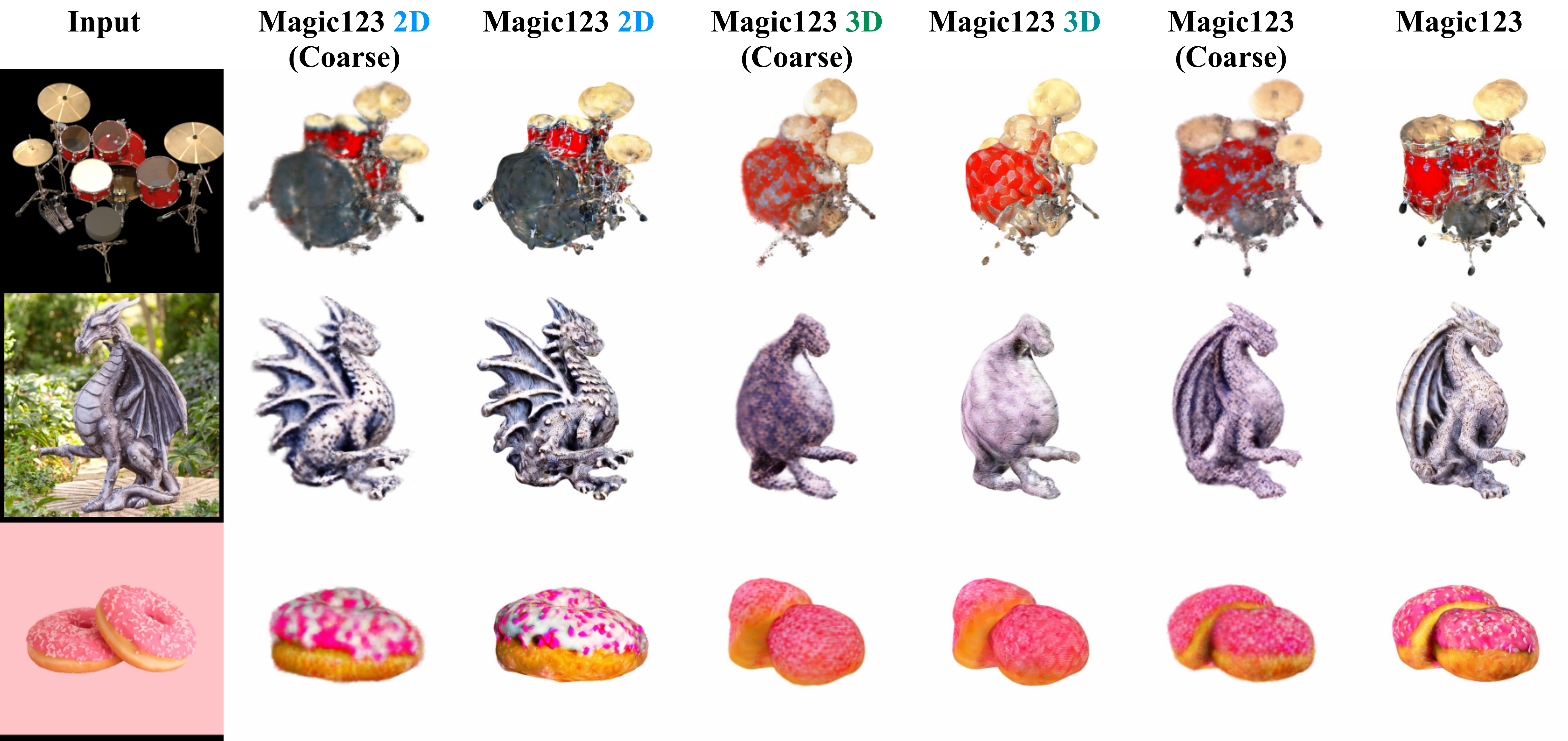}
\caption{\textbf{Ablation study (qualitative).} We qualitatively compare the novel view renderings from the coarse and fine stages in Magic123. We ablate utilizing only 2D prior ($\lambda_{2D}$=$1$,$\lambda_{3D}$=$0$), utilizing only 3D prior ($\lambda_{2D}$=$0$,$\lambda_{3D}$=$40$), and utilizing both 2D and 3D priors ($\lambda_{2D}$=$1$,$\lambda_{3D}$=$40$).
}
\label{fig:2d-3d-coarse-fine}
\end{figure}

\begin{figure}[t] 
\centering
\includegraphics[width=\textwidth]{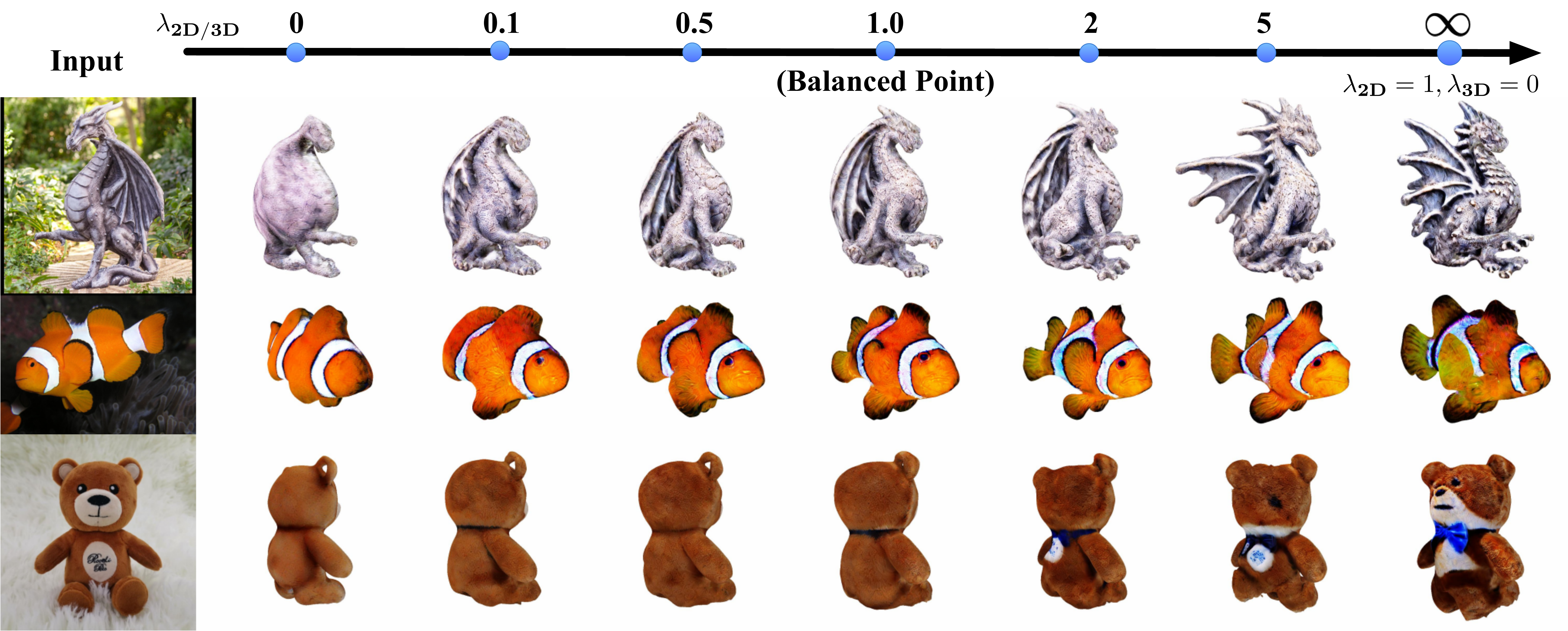}
\caption{\textbf{Setting $\lambda_{2D/3D}$}. We study the effects of $\lambda_{2D/3D}$ on Magic123. Increasing $\lambda_{2D/3D}$ leads to a 3D geometry with higher imagination and less precision and vice versa. $\lambda_{2D/3D}$=$1$ provides a good balance and thus is used as the default. 
}
\label{fig:lambda_2d-3d}
\end{figure}
\section{Related work}\label{sec:related}

\inlinesection{Multi-view 3D reconstruction}
Multi-view 3D reconstruction aims to recover the 3D structure of a scene from its 2D RGB images captured from different camera positions~\cite{UncalibratedStereoRig, RomeInADay}.
Classical approaches usually recover a scene's geometry as a point cloud using SIFT-based~\cite{SIFT} point matching~\cite{colmap_sfm, colmap_mvs}.
More recent methods enhance them by relying on neural networks for feature extraction (\eg \cite{mvsnet, DeepMVS, rmvsnet, fastmvsnet}).
The development of Neural Radiance Fields (NeRF)~\cite{NeRF, NeuralVolumes} has prompted a shift towards reconstructing 3D as volume radiance~\cite{VolumeRenderingDigest}, enabling the synthesis of photo-realistic novel views~\cite{RefNeRF, MipNeRF, MipNeRF-360}.
Subsequent works have also explored the optimization of NeRF in few-shot (\eg \cite{DietNeRF, InfoNeRF, WideBaseline}) and one-shot (\eg \cite{PixelNeRF, EG3D}) settings.
NeRF does not store any 3D geometry explicitly (only the density field), and several works propose to use a signed distance function to recover a scene's surface~\cite{IDR, NeuS, VolSDF, HF-NeuS, PatchNeuS}, including in the few-shot setting as well (\eg \cite{MonoSDF, NeRS}).

\inlinesection{In-domain single-view 3D reconstruction}
3D reconstruction from a single view requires strong priors on the object geometry since even epipolar constraints~\cite{MultiViewGeometry} cannot be imposed in such a setup.
Direct supervision in the form of 3D shapes or keypoints is a robust way to impose such constraints for a particular domain, like human faces~\cite{3DMM, 3DMM-learnt}, heads~\cite{FLAME, FaceVerse}, hands~\cite{ExpressiveBodyCapture} or full bodies~\cite{SMPL, 3D-pose-baseline}.
Such supervision requires expensive 3D annotations and manual 3D prior creation. Thus several works explore unsupervised learning of 3D geometry from object-centric datasets (\eg \cite{CMR, TARS-3D, InternetMM, TotalMoving3DFace, UCMR, CMS, UVA}).
These methods are typically structured as auto-encoders~\cite{Unsup3D, CategoryShapes, SDF-SRN} or generators~\cite{pix2nerf, IDE-3D} with explicit 3D decomposition under the hood. Due to the lack of large-scale 3D data, these methods are limited to simple shapes (\eg chairs, cars) and cannot generalize to more complex or uncommon objects (\eg dragons, statues).

\inlinesection{Zero-shot single-view 3D reconstruction}
Foundational multi-modal networks~\cite{CLIP, DINO, LDM} have enabled various zero-shot 3D synthesis tasks.
Earlier works employed CLIP~\cite{CLIP} guidance for 3D generation~\cite{DreamFields, AvatarCLIP, CLIP-Mesh, Dream3D}
% , stylization~\cite{text2mesh, 3DHighlighter, NeRF-Art}
and manipulation~\cite{StyleCLIP, DFF, TextDeformer} from text prompts.
Modern zero-shot text-to-image generators~\cite{DALL-E, LDM, DALLE-2, Imagen, ediffi, Make-a-Scene} allowed to improve these results by providing stronger synthesis priors~\cite{DreamFusion, SJC, Latent-NeRF, text2tex, sdfusion}.
DreamFusion~\cite{DreamFusion} is a seminal work that proposed to distill an off-the-shelf diffusion model~\cite{Imagen} into a NeRF~\cite{NeRF, MipNeRF-360} for a given text query.
It sparked numerous follow-up approaches for text-to-3D synthesis (\eg \cite{Magic3D, Fantasia3D}) and image-to-3D reconstruction (\eg \cite{DITTO-NeRF, RealFusion, Zero-1-to-3, Make-It-3D}).
The latter is achieved via additional reconstruction losses on the frontal camera position~\cite{Zero-1-to-3} and/or subject-driven diffusion guidance~\cite{DreamBooth3D, Magic3D}.
The developed methods improved the underlying 3D representation~\cite{Magic3D, Fantasia3D, stable-dreamfusion} and 3D consistency of the supervision~\cite{Zero-1-to-3, 3DFuse}; explored task-specific priors~\cite{Text2Room, Farm3D, TEXTure} and additional controls~\cite{SKED}.
Similar to the recent image-to-3D generators~\cite{Zero-1-to-3, RealFusion}, we also follow the DreamFusion~\cite{DreamFusion} pipeline, but focus on reconstructing a high-resolution, textured 3D mesh using a joint 2D and 3D priors.  

\section{Conclusion and discussion}\label{sec:con}

% \inlinesection{Conclusion}\label{sec:conclusion}
This work presents Magic123, a two-stage coarse-to-fine solution for generating high-quality, textured 3D meshes from a \emph{single} unposed image. By leveraging both 2D and 3D priors, our approach overcomes the limitations of existing studies and achieves state-of-the-art results in image-to-3D reconstruction. The trade-off parameter between the 2D and 3D priors allows for control over the balance between exploration and exploitation of the generated geometry. Our method outperforms previous techniques in terms of both realism and level of detail, as demonstrated through extensive experiments on real-world images and synthetic benchmarks. Our findings contribute to narrowing the gap between human abilities in 3D reasoning and those of machines, and pave the way for future advancements in single image 3D reconstruction.
The availability of our code, models, and generated 3D assets will further facilitate research and applications in this field.

\inlinesection{Limitation}\label{sec:limit}
One of the limitations is that we assume the reference image is taken from the front view. This assumption leads to poor geometry when the reference image does not conform to the front-view assumption, \eg a photo of a dish on the table taken from the up view. Our method will instead focus on generating the bottom of the dish and table instead of the dish geometry itself. This limitation can be alleviated by a manual reference camera pose tuning or camera estimation. Another limitation of our work is the dependency on the preprocessed segmentation \cite{DPT} and the monocular depth estimation model \cite{MiDaS}. Any error on these modules will creep into the later stages and affect the overall generation quality. 
Similar to previous work, Magic123 also tends to generate over-saturated textures due to the usage of the SDS loss. The over-saturation issue becomes more severe for the second stage because of the higher resolution. 
% See \supp for more details about the limitations.
% \todo{another failure case, should be related to when MiDas or the segmentation fails.}

% \mysection{Acknowledgement}
\inlinesection{Acknowledgement}\label{sec:ack}
The authors would like to thank Xiaoyu Xiang for the insightful discussion and Dai-Jie Wu for sharing Point-E and Shap-E results. This work was supported by the KAUST Office of Sponsored Research through the Visual Computing Center funding, as well as, the SDAIA-KAUST Center of Excellence in Data Science and Artificial Intelligence (SDAIA-KAUST AI). Part of the support is also coming from KAUST Ibn Rushd Postdoc Fellowship program.

\clearpage \clearpage 
{\small
\bibliographystyle{plain}
\bibliography{main}
}

\end{document}